# SLSG: Industrial Image Anomaly Detection by Learning Better Feature Embeddings and One-Class Classification

Minghui Yang, Jing Liu*, Zhiwei Yang, and Zhaoyang Wu

*Guangzhou Institute of Technology, Xidian University, Guangzhou 510555, China.*

## Abstract

Industrial image anomaly detection under the setting of one-class classification has significant practical value. However, most existing models struggle to extract separable feature representations when performing feature embedding and struggle to build compact descriptions of normal features when performing one-class classification. One direct consequence of this is that most models perform poorly in detecting logical anomalies which violate contextual relationships. Focusing on more effective and comprehensive anomaly detection, we propose a network based on self-supervised learning and self-attentive graph convolution (SLSG) for anomaly detection. SLSG uses a generative pre-training network to assist the encoder in learning the embedding of normal patterns and the reasoning of position relationships. Subsequently, SLSG introduces the pseudo-prior knowledge of anomaly through simulated abnormal samples. By comparing the simulated anomalies, SLSG can better summarize the normal features and narrow down the hypersphere used for one-class classification. In addition, with the construction of a more general graph structure, SLSG comprehensively models the dense and sparse relationships among elements in the image, which further strengthens the detection of logical anomalies. Extensive experiments on benchmark datasets show that SLSG achieves superior anomaly detection performance, demonstrating the effectiveness of our method.

## 1. Introduction

Surface defect or anomaly detection of industrial images refers to identifying heterogeneous or unexpected patterns in images, it is a classification task that identifies normal and anomaly. However, collecting comprehensive abnormal images and using supervised methods for surface defect detection is extremely challenging, as abnormal samples in practical applications are rare and abnormal patterns are various. Therefore, the setting of one-class learning using only normal samples for model training, i.e. the semi-supervised learning paradigm, is better adapted to anomaly detection tasks.

Recently, many methods have been proposed to detect whether the new input image matches the distribution of the normal patterns to complete anomaly detection. To perform feature embedding on normal patterns, reconstruction-based methods [1,2,3,4,5] attempt to train a deep neural network that can only reconstruct normal features. Such methods expect features corresponding to abnormal regions in the image to have larger reconstruction differences. However, the contradiction between the reconstruction and generalization capabilities of deep learning does not satisfy this expectation well [6]. Besides, feature representation-based methods [7,8,9,10] attempt to perform feature embedding on normal patterns using an encoder pre-trained on the ImageNet dataset [11]. Although the accuracy of such methods for anomaly detection is usually high, the speed of detection is generally slow. Moreover, using an encoder pre-trained on the ImageNet for feature embedding also has some shortcomings: (1) The encoder pre-trained on the source domain dataset (ImageNet) does not guarantee that the features of normal and abnormal patterns extracted from the target domain dataset are distinguishable [12,13]; (2) The pre-training process based on the ImageNet does not explicitly model the position relationships [14]; (3) The pre-training process based on the ImageNet generally completes the image classification task. Under the supervision of the image-level classification task, the granularity of the extracted features may be large, so it is not suitable for fine-grained pixel-level anomaly judgments.

In addition, the semi-supervised image anomaly detection models are trained under the constraint of one-class classification and compress all normal samples into a hypersphere. However, most models observe only normal samples during training and without additional constraints, which makes it difficult to establish a compact description of normal samples [15,16]. In this case, the radius of the hypersphere used for one-class classification is usually too large or even infinite.

---

*Corresponding author.

More importantly, most existing semi-supervised methods based on convolutional neural networks (CNNs) only have good detection performance for local structural anomalies, such as scratches and stains. But they have poor detection performance for logical anomalies, which violate logical constraints [17], such as missing and misplaced elements. The detection of logical anomalies requires the model to fully perceive the position information using large receptive fields during the process of performing both feature embedding and one-class classification, which is difficult for CNNs [18].

In summary, existing semi-supervised anomaly detection methods face challenges in performing more comprehensive feature embedding and better one-class classification. To address these challenges, in this paper, we propose a network based on self-supervised learning and self-attentive graph convolution (SLSG) to accomplish semi-supervised image anomaly detection.

First, to learn better feature embedding, we use a generative pre-training network (GPT-Net) to pre-train the encoder. GPT-Net overlays a large-scale mask on the input industrial image, and then performs the pixel-level self-supervised task of mask inpainting. Using anomaly-specific datasets for encoder pre-training alleviates the problem of domain adaptation faced by feature representation-based methods. Also, by reasoning about the possible forms of the masked regions, the encoder can better perceive the position information and capture long-range dependencies in images.

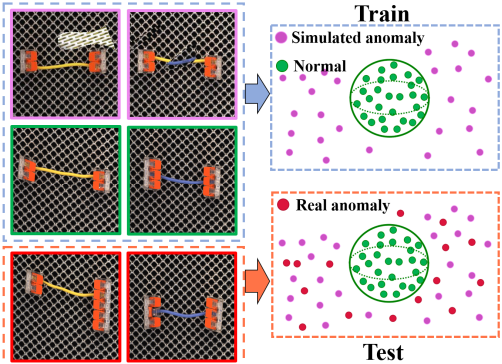

Figure 1. SLSG uses normal and simulated anomalies for training and completes anomaly discrimination directly during the inference stage.

Then, to learn better one-class classification, as shown in Fig. 1, we introduce pseudo-prior knowledge of the anomaly through simulated abnormal images during the training stage to add the training constraints. By comparing and distinguishing simulated anomalies, SLSG can better summarize what is normal and shrink the hypersphere used to describe the normal features. With the learned knowledge, SLSG can directly generalize to the detection of real anomalies in the inference stage. To ensure the generalization performance of SLSG, there are two key points to note: the radius of the hypersphere being small enough and the data distribution of the simulated abnormal samples being widespread enough. The high consistency of normal samples in the same industrial production line ensures that the distribution of normal features is concentrated, thus the first key point can be guaranteed. To ensure the second key point, we propose an efficient anomaly simulation strategy that simulates comprehensive abnormal samples from the perspective of structural and logical anomalies.

Furthermore, to make the hypersphere used for one-class classification better cover the normal logical relationships, we introduce a self-attention-based graph convolutional network (SG block). The SG block constructs a more general graph by eliminating locally redundant semantic similarities. And the update process of the nodes in the graph fuses global extensive information and cross-neighborhood key information, which enhances the ability of SLSG to detect logical anomalies.

With extensive validation on multiple datasets, SLSG achieves precise anomaly detection and reaches state-of-the-art (SOTA) performance with 90.3% ROC-AUC value on the MVTec LOCO AD dataset [17], which comprehensively covers both structural and logical anomalies. Moreover, the end-to-end fully convolutional structure of SLSG ensures the real-time performance in the inference stage, which can process 54 images per second using an NVIDIA GTX 1080Ti GPU.

The main contributions of this paper are summarized as follows:
● Based on the generative pre-training network, we design a self-supervised task of mask inpainting for encoder pre-training, alleviating the domain adaptation problem of transfer learning and enhancing the sensitivity of the encoder to position information;
● We propose an anomaly simulation strategy from two aspects of structural and logical anomalies, which ensures the diversity of simulated abnormal samples and enhances the generalization capability of the model;
● We propose a self-attention-based graph convolutional network to model global extensive relationships and cross-neighborhood key relationships among nodes to help the model perceive the contextual information in images;
● Extensive experiments demonstrate the SOTA performance and faster speed of SLSG in detecting structural and logical anomalies.

## 2. Related Work

### 2.1 Image Surface Defect Detection

Reconstructing the input image or high-level features of the image is a traditional method for semi-supervised image surface defect detection. This type of method mainly uses autoencoders [1,2], generative adversarial networks [3,4,5], or student-teacher networks [19,20,21,22,23] for reconstruction and performs anomaly discrimination based on the reconstruction differences. Although experience only normal samples during training, the models may also reconstruct abnormal regions correctly during the inference stage due to the excessive generalization capabilities of deep neural networks, and thus the discriminations based on the reconstruction differences will fail. To limit the generalization abilities of the reconstruction models, Deng et al. [23] proposed a reverse distillation strategy, Zavrtanik et al. [24] and Pirnay et al. [25] performed a random mask on the input image, and Gong et al. [26] introduced a memory module. However, there has been no effective method to balance the reconstruction ability and generalization ability of deep neural networks well, thus the performance of reconstruction-based models leaves much space to improve. SLSG is based on the autoencoder, but its anomaly judgments are output end-to-end, rather than obtained indirectly through the reconstruction differences.

Recently, feature representation-based methods have been demonstrated to have good anomaly detection performance [7,8,9]. This type of method first performs feature embedding on all samples in the training set, and then matches the minimum distance of features between the test and training samples during the inference stage. Further, they perform anomaly judgments based on the minimum distance. Although no complex training process is required, such methods generally require traversing a large number of normal features to complete the feature matching during the inference stage, which affects the inference speed of the model application.

## 2.2 Self-Supervised Learning in Surface Defect Detection

Self-supervised learning has significantly improved the performance of models in both computer vision [27,28] and natural language processing [29,30]. Apart from tasks such as reconstructing images or features, there are many self-supervised tasks for image anomaly detection. From the perspective of prediction, SSPCAB [31] proposed a plug-and-play self-supervised predictive convolutional attentive block for anomaly detection. Around the perspective of simulating abnormal samples, CutPaste [32] and AnoSeg [33] randomly copied and pasted rectangular regions within the normal image to generate structural anomalies. DRAEM [34] pasted additional texture noise into the image to generate texture anomalies. However, to complete the pixel-level anomaly localization, AnoSeg and DRAEM still had to perform the auxiliary operation of de-anomaly. CutPaste had to use methods such as Grad-CAM [35] to achieve anomaly localization indirectly. In addition, existing anomaly simulation strategies only consider the case of structural anomalies, and the incomplete pretext task is not conducive to the model distinguishing between normal samples and logically abnormal samples. More effectively, SLSG simulates comprehensive abnormal samples and uses simulated anomalies to directly achieve end-to-end pixel-level anomaly localization.

## 2.3 Position Modeling in Surface Defect Detection

For the task of image anomaly detection, the vast majority of existing datasets and methods do not specifically consider logical anomalies in images. Since CNN is insensitive to position information [18], directly applying existing methods to detect logical anomalies is less effective. In order to improve the abilities of models to perceive the logical relationships in normal patterns, special module designs are required, but there are few such methods at present. GCAD [17] introduced the global and local branches using student-teacher networks [36]. The global branch mainly modeled the semantic information of logical anomalies using large receptive fields, while the local branch mainly modeled the visual information of structural anomalies. Finally, the results of the two branches are fused to complete the comprehensive detection. AnoSeg [33] introduced the position information from the perspective of data input. It concatenated the RGB channel of the original image with an additional coordinate channel as the model input. To model position information in images, we use a self-supervised task to learn the reasoning of position relationships and use the graph convolutional network (GCN) to capture across-neighborhood position relationships.

## 3. Method

As shown in Fig. 2, SLSG consists of a generative pre-training network (GPT-Net) and an anomaly segmentation network (SegNet). Specifically, GPT-Net completes the pre-training of the encoder using a self-supervised task of mask inpainting, so as to help the encoder learn better feature embeddings (Section 3.1). With the pre-trained encoder, SegNet performs the self-supervised task of anomaly segmentation by introducing simulated abnormal samples (Section 3.2). By contrasting simulated anomalies, the decoder of SegNet, i.e. the one-class classifier, can enhance its ability to summarize normal features. Moreover, we introduce a self-attention-based graph convolutional network (SG block) in the decoding process of SegNet, which further improves the performance of SegNet to detect logical anomalies (Section 3.3). In the inference stage, using the learned one-class classification decision boundary, SegNet can easily generalize to anomaly detection in real scenes and directly complete the pixel-level anomaly segmentation. In this section, we will describe the key parts of SLSG in detail.

## 3.1 Learning Feature Embeddings using GPT-Net

To achieve anomaly detection, the feature representation-based methods use an encoder that is pre-trained with the ImageNet dataset to extract the deep features of the industrial images. However, as described in Section 1, pre-training with the ImageNet dataset has the disadvantages of poor domain adaptation and being unable to effectively model position information. To perform more comprehensive feature embedding, we design the GPT-Net based on anomaly-specific datasets and masked autoencoder [37] to pre-train the encoder.

First, GPT-Net uniformly divides the input image into $S \times S$ patches based on the grid structure and randomly masks $p\%$ of the patches. Subsequently, the masked image is fed into GPT-Net and the image reconstruction task is completed by the demasking operation. In the process of repairing these grid masks, GPT-Net can learn the relative position relationships among elements in normal images within small neighborhoods. In addition, to improve the ability of GPT-Net to model remote relationships, on the basis of the grid masks, we superimpose the large masks with random positions and larger scales. By reasoning about the long-range relationships, the encoder is forced to aggregate the logical information present in the image with a larger effective receptive field.

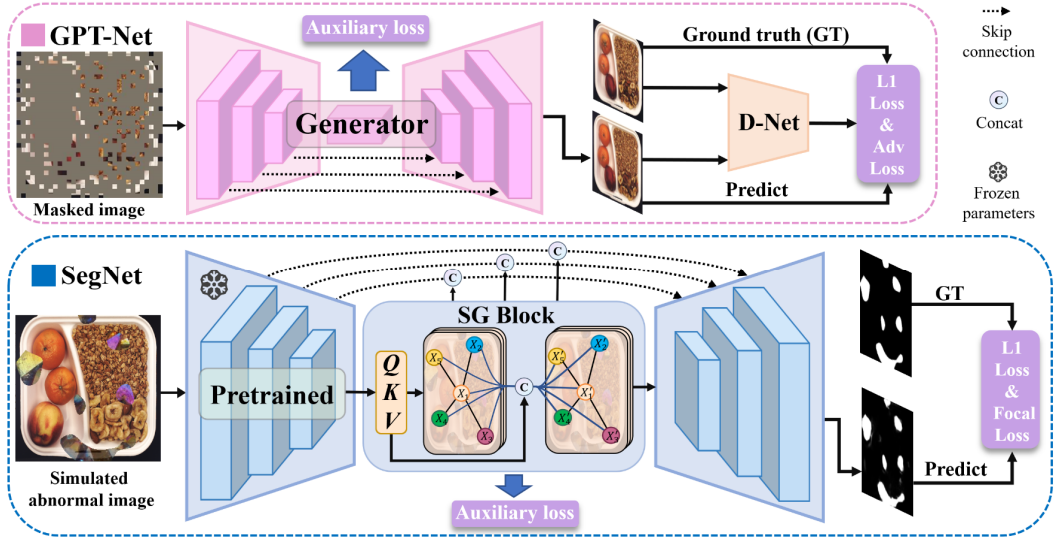

Figure 2. The pipeline of SLSG. SLSG uses a generative pre-training network (GPT-Net) to pre-train the encoder, and this process introduces adversarial loss through the discriminative network (D-Net) to improve the sharpness of the reconstructed image. Then, SLSG uses the pre-trained encoder and the one-class classification decoder to compose the segmentation network (SegNet). SegNet performs end-to-end pixel-level anomaly localization with the help of simulated abnormal samples. Also, SegNet introduces a self-attention-based graph convolutional network (SG block) at the bottleneck structure to better capture logical anomalies.

From the perspective of network structure, we effectively design the GPT-Net from the following aspects:

(1) After the original image has been masked, its data distribution has changed considerably. Hence, it is necessary to boost the model capacity by adding skip connections [38] to improve the reconstruction ability of GPT-Net. Using skip connections, GPT-Net can fully use the visual and semantic information of different depths of the encoder;

(2) Due to the presence of unmasked regions in the input image, the addition of skip connections may cause GPT-Net to focus excessively on the shallow visual information. Therefore, while the decoder completes the image reconstruction, GPT-Net also uses the features at the bottleneck structure to complete the image reconstruction directly. Specifically, GPT-Net introduces an auxiliary loss [39] at the bottleneck structure to emphasize the training of the deeper part of the encoder;

(3) Although the images acquired in the same industrial pipeline are highly consistent, there still exist some differences in normal patterns, especially in some textural areas. If the loss function simply measures the pixel-by-pixel distance between the reconstructed image of GPT-Net and the real image, such as $\ell 1$ loss, the reconstructed image will suffer from blurring. Considering the impressive performance of the generative adversarial network (GAN) [40] in super-resolution reconstruction [41,42], we introduce the adversarial loss through a Markov discriminator [43] D-Net, i.e., the GPT-Net can be seen as a generator. The generator improves the sharpness of the reconstructed image by combining $\ell 1$ loss and adversarial loss to further enhance the feature representation ability of the encoder.

It should be noted that the role of GPT-Net is to train an encoder that is used to model normal patterns including complex logical relationships. This process does not take into account any prior knowledge of anomalies, as it is difficult for the encoder to build a consistent description of the variable abnormal patterns.

### 3.2. Anomaly Simulation Strategy

For image anomaly detection within the industrial context, strict quality control of productions leads to a relative lack of abnormal samples. Meanwhile, the abnormal samples also appear in various forms, making it difficult to use a supervised learning framework for anomaly detection. However, if only simply observing normal samples under the semi-supervised constraints, the model may consider any inputs to be normal. In this case, the decision boundary learned by the one-class classifier is meaningless [15]. To avoid this possible collapse, we introduce the simulated abnormal samples during the training of SegNet. The decoder of SegNet can build a compact description of normal patterns by comparing a wide range of simulated anomalies. To generate simulated anomaly samples with a wider distribution, building on our previous work [16], we design a more comprehensive anomaly simulation strategy that considers both structural and logical types of anomalies.

As shown in Fig. 3, the main steps of SegNet to perform anomaly simulation are as follows: SegNet first calculates the Hadamard product of the binarized Perlin noise [44] and the binarized target foreground to obtain the mask image $M$. In the process of generating $M$, the introduction of the binarized target foreground can ensure that simulated anomalies will not appear in the meaningless background areas of images, which we call the foreground enhancement strategy. After that, the mask image $M$ extracts the anomaly foreground image in the noise source image $I_N$. Then, some traditional data augmentation operations are performed on the anomaly foreground image, such as random adjustments of brightness and contrast. Finally, we superimpose the anomaly foreground image onto the normal image $I$ to generate the simulated abnormal image $I_A$. As shown in Eq. 1, this process introduces a transparency factor $\delta$ to make the generated simulated abnormal image more realistic [34]:

$$I_A = \bar{M} \odot I + M \odot (\delta \times I_N + (1-\delta) \times I) \quad (1)$$

where $\bar{M}$ is the image obtained by inverting $M$.

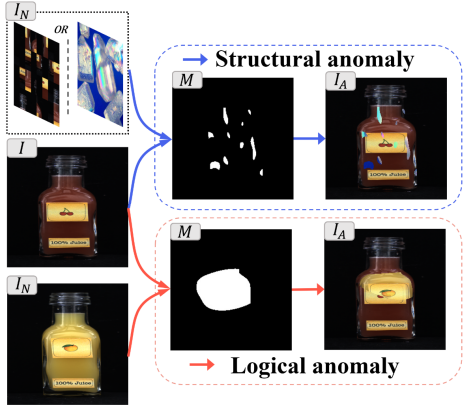

Figure 3. Structural anomaly simulation and logical anomaly simulation. The Hadamard product of the mask image $M$ and the noise source image $I_N$ defines the anomaly foreground. The anomaly foreground is superimposed on the normal image $I$ to produce the simulated anomaly image $I_A$.

Considering structural and logical anomalies, the anomaly simulation strategy of SegNet is divided into two branches, namely structural anomaly simulation and logical anomaly simulation. In the branch of structural anomaly simulation, the noise source image $I_N$ comes from two data sources. One is the DTD texture dataset [45], which is designed to simulate heterologous structural anomalies. And the other is the image obtained by randomly permuting the normal image $I$, which is designed to simulate homologous structural anomalies. In the branch of logical anomaly simulation, after excluding the current input image $I$, we randomly select a normal image in the training set as $I_N$, which will also undergo a random rotation for data augmentation when used. In addition, the area of logical anomalies in real scenarios is generally large, and reducing the data distribution differences between simulated and real anomalies can facilitate the anomaly detection. Therefore, during the process of logical anomaly simulation, SegNet makes the foreground region of the generated mask image $M$ more concentrated by controlling the parameters of Perlin noise and the binarization threshold.

Using normal samples and simulated abnormal samples, SegNet builds a self-supervised semantic segmentation task. At the same time, combined with the highly consistent normal pattern in the same industrial pipeline, SegNet can better learn what is normal by comparing the pseudo-prior knowledge of anomalies, and can easily generalize to the detection of real anomalies during the inference stage.

### 3.3 Self-Attention-Based GCN

Generally, the detection of logical anomalies is relatively difficult compared to structural anomalies, as logical anomalies usually need to be judged with the assistance of large receptive fields and cross-neighborhood position relationships. After the pre-training with GPT-Net, the encoder of SegNet can extract features containing position information. Meanwhile, enhancing the sensitivity of the decoder to position relationships can further improve the ability of the model to detect logical anomalies. Therefore, we introduce the SG block in the decoding process of SegNet. SG block uses CNN-extracted deep feature maps for modeling, as the deep feature maps have more semantic information while ensuring a larger theoretical perceptual field. Further, considering that the CNN architecture is not good at modeling remote dependencies among patches in deep feature maps, as shown in Fig. 4, the SG block uses a self-attention-based graph convolutional network to model global dense relationships and cross-neighborhood sparse relationships.

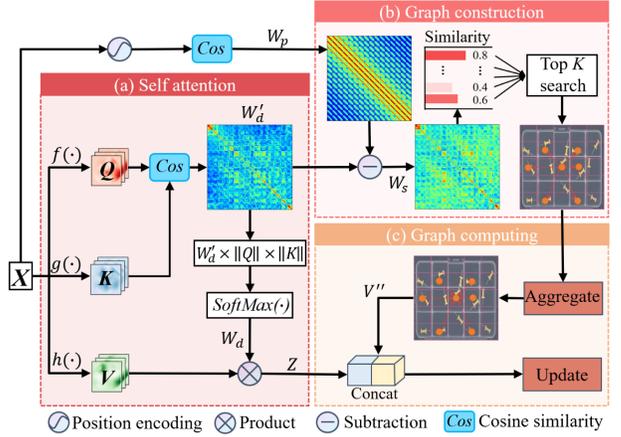

Figure 4. Self-attention-based graph convolutional block.

**Dense Relationships.** For the deep feature map of the encoder output $X \in \mathbb{R}^{W \times H \times C_{in}}$, applying self-attention [46,47] to model the dense relationships among all patches at the global scale obeys the following equations:

$$W_d = Softmax\left(\frac{f(X)g(X)^T}{\sqrt{d}}\right) \quad (2)$$

$$Z = W_d h(X) \quad (3)$$

In Eq. 2, we first uses two functions $f(\cdot)$ and $g(\cdot)$ to perform feature conversion and dimension adjustment on $X$ to obtain $f(X), g(X) \in \mathbb{R}^{N \times \frac{C_{in}}{2}}$, where $N = W \times H$. $f(X)$ and $g(X)$ correspond to the query and key respectively in the self-attention mechanism. Then, the product of query and key is processed by the $softmax(\cdot)$ to get the dense similarity matrix $W_d \in \mathbb{R}^{N \times N}$ of the global space. The dense similarity matrix $W_d$ can query the similarity between any two patches in feature $X$. Subsequently, the dense similarity matrix $W_d$ is used to weight and sum the feature $h(X) \in \mathbb{R}^{N \times C_{out}}$ to obtain the output of the self-attention network $Z \in \mathbb{R}^{N \times C_{out}}$, and by expanding $Z$ into a two-dimensional feature map, we can get $Z' \in \mathbb{R}^{W \times H \times C_{out}}$. The functions $f(\cdot)$, $g(\cdot)$, and $h(\cdot)$ in Eq. 2 and 3 are implemented using a convolution layer with 1×1 kernel, and $C_{in}=C_{out}$ in the specific implementation.

**Sparse Relationships.** For the deep feature map of the encoder output $X \in \mathbb{R}^{W \times H \times C_{in}}$, its grid structure can be regarded as a special case of the graph. In this graph, each patch with channel $C_{in}$ can be regarded as a node, and edges only exist between the node and its surrounding neighborhoods. In the previous discussion, through self-attention, we explored the dense relationships among all nodes. Further, constructing a more generalized graph and thus using GCN to emphasize across-neighborhood key relationships among nodes may help the model handle more complex logical anomalies.

For $X \in \mathbb{R}^{W \times H \times C_{in}}$, we first convert it to a more general graph structure $G(V, E)$, where $V$ is the set of nodes, $V = \{v_i\}_{i=1:HW}$ and $v_i \in \mathbb{R}^{C_{in}}$. For the construction of edge set $E$, we can use the similarity matrix $W_s$ to index the top-K similar nodes $N_i = \{v_i^j\}_{j=1:K}$ to each node $v_i$ and establish the connection relationships [48], i.e., $E = \{e_{ij}\}_{i=1:HW, j=1:K}$. So, how to construct the similarity matrix $W_s$?

Using the dense similarity matrix $W_d$ generated by self-attention directly to build sparse edges among nodes suffers from certain drawbacks. Due to the spatial continuity of semantic information, for a node $v_i$ in the graph, its most similar nodes indexed by $W_d$ have a high probability of appearing in the surrounding neighborhood of node $v_i$. In this case, the purpose of modeling the remote dependencies with the help of GCN is not achieved. Hence, we need to eliminate the redundant semantic similarity due to the proximity of positions between nodes. First, we perform a two-dimensional sine-cosine position encoding [49] for $X$ to obtain a spatial feature map $P \in \mathbb{R}^{N \times C_p}$, where $C_p$ is the dimension of the position encoding. Then we calculate the cosine similarity between $P$ and $P^T$ according to Eq. 4 to get the positional similarity matrix $W_p \in \mathbb{R}^{N \times N}$. Further, as shown in Eq. 5, we recompute the dense similarity matrix $W_d'$ using the cosine similarity, and subtract the positional similarity matrix from $W_d'$ to obtain the sparse similarity matrix $W_s \in \mathbb{R}^{N \times N}$. $W_s$ is used to construct the graph structure. By eliminating the position similarity, $W_s$ puts more emphasis on the semantic similarity between across-neighborhood nodes.

$$W_p = CosSim(P, P^T) \quad (4)$$
$$W_s = CosSim(f(X), g(X)^T) - W_p \quad (5)$$

After completing the construction of the graph $G(V, E)$, we perform the feature aggregation of the nodes using max-relative graph convolution [50]:

$$v_i' = max\{v_i^j - v_i\}, j = 1:K \quad (6)$$

by adjusting the dimension of the aggregated node set $V' = \{v_i'\}_{i=1:HW}$, a two-dimensional feature map $V'' \in \mathbb{R}^{W \times H \times C_{in}}$ can be obtained.

Through the Eq. 3 and 6, SG block gets the feature $Z'$ containing global extensive information and the feature $V''$ containing cross-neighborhood key information, respectively. We concatenate $Z'$ and $V''$ in the channel dimension, and get the output of the SG block $X' \in \mathbb{R}^{W \times H \times C_{out}}$ through the update function $u(\cdot)$:

$$X' = u([Z', V'']) \quad (7)$$

where the update function $u(\cdot)$ is also implemented by a convolution layer with 1×1 kernel.

Compared to $X$, $X'$ has larger receptive field, richer semantic information, and greater ability to represent the complex logical relationships among individual patches in an image. $X'$ will be used as the input of the decoder of SegNet to complete the pixel-level anomaly localization. At the same time, we also feed $X'$ into blocks of different depths in the decoder via skip connections to further emphasize the effect of $X'$, making all parts of the decoder focus on the semantic information and the logical relationships among patches as much as possible. With $X'$, the decoder of SegNet can better optimize the decision boundary for one-class classification. Using this decision boundary, in the inference stage, SegNet can perform anomaly localization directly after acquiring the input query.

### 3.4 Optimization Objectives

To improve the diversity and clarity of the reconstructed images, SLSG uses GPT-Net as a generator and introduces a fully convolutional Markov discriminator to form a GAN. SLSG uses least square GAN loss [51] to complete the optimization of the GAN, which ensures the GAN converge more stably while generating images with higher quality. Specifically, we optimize the adversarial loss between the generator $G$ and the discriminator $D$ using Eq. 8 and 9:

$$L_G = \frac{1}{2}\left(D(G(x_m)) - 1\right)^2 \quad (8)$$
$$L_D = \frac{1}{2}\left(D(G(x_m))\right)^2 + \frac{1}{2}(D(x) - 1)^2 \quad (9)$$

where $x_m$ is the input of the generator, i.e. the masked image, and $x$ is the original image.

Further, to reduce the difference between reconstructed image and original image, we additionally optimize the generator using $\ell 1$ loss. Also, as explained in Section 3.1, we use the $\ell 2$ distance-based auxiliary loss $L_{aux}^R$ at the bottleneck structure of the GPT-Net to optimize the extraction of deep features. Finally, the optimization objective of the GPT-Net is:

$$L_R = \alpha_1 L_G + \alpha_2 L_{\ell 1} + \alpha_3 L_{aux}^R \quad (10)$$

For SegNet, SLSG uses both the $\ell 1$ loss $L_{\ell 1}$ and the focal loss $L_f$ [52] to optimize the model parameters. The $\ell 1$ loss can alleviate the blurring problem of the predicted segmented image. The focal loss can alleviate the pixel-level data imbalance problem. Similarly, combined with the auxiliary loss $L_{aux}^S$ of SegNet at the bottleneck structure, we get the optimization objective of SegNet:

$$L_S = \alpha_4 L_f + \alpha_5 L_{\ell 1} + \alpha_6 L_{aux}^S \quad (11)$$

where $\alpha$ is the balance factor among the different loss functions.

## 4. Experiments

In this section, we experimentally compare the detection accuracy and speed of SLSG with other models on three benchmark datasets, MVTec AD [53], BeanTech AD [54], and MVTec LOCO AD. Meanwhile, we also evaluate the functionality of different modules in SLSG to further demonstrate the effectiveness and working mechanism of SLSG.

### 4.1. Datasets and Evaluation Metric

The MVTec AD and the BeanTech dataset are benchmark datasets for semi-supervised image surface defect detection. The training set of MVTec AD contains 15 categories of approximately 3600 normal images. The test set contains normal images and 73 kinds of abnormal images in real scenes, and the type, position, and size of the abnormal regions are random. The BeanTech dataset follows the setting of one-class learning as does the MVTec AD dataset, but its training set contains only 3 categories of approximately 1800 images. For the evaluation of image-level anomaly detection and pixel-level anomaly localization of these two datasets, consistent with [1,2,3,7,8], we use ROC-AUC as the evaluation metric.

The vast majority of anomalies in the MVTec AD dataset and the BeanTech dataset are structural anomalies, and we use the MVTec LOCO AD dataset to measure the ability of the model to detect logical anomalies. The LOCO AD dataset contains 5 categories of approximately 3644 images in industrial scenarios, and evenly covers both structural and logical types of anomalies. Considering the ambiguity of the logical anomaly determination at the pixel level, GCAD [17] defines the saturated per-region overlap (sPRO) to evaluate the performance of the model for anomaly localization. Therefore, for the evaluation of pixel-level anomaly localization on the LOCO AD datasets, consistent with [17], we calculate the area under the FPR-sPRO curve up to the false positive rate is 5%, then normalize it to obtain the evaluation metric sPRO-AUC with a score between 0 and 1. And for the image-level evaluation metric, we still use the ROC-AUC.

### 4.2. Implementation Details

For GPT-Net, we resize the input image to 256×256, divide the image into uniform patches using grids of 8×8, and randomly mask 80% of the patches. In addition, on the basis of the grid patch, we generate large masks with lengths and widths between 40-150. To ensure the inference efficiency in industrial scenarios, GPT-Net uses resnet18 [55] as the encoder. The decoder is a symmetrical structure with the encoder, containing four stacked up-sampling blocks and Conv blocks. The up-sampling block includes an up-sampling layer and a convolution group which is made up of convolution layer, batch normalization, and ReLU layer. The Conv block includes two stacked convolution groups. The discriminator D-Net consists of four convolution groups and outputs a single-channel feature map for the judgments of real or false. GPT-Net uses the Adam optimizer, iterates 40 epochs, and the batch size is set to 6. We utilize grid search for the optimization of hyper-parameters: the learning rate for the generator is 0.001 and for the discriminator is 0.0006; $\alpha_1$, $\alpha_2$, and $\alpha_3$ in the loss function of generator is set to 1, 0.8, and 1, respectively.

SegNet also uses resnet18 as the encoder and the structure of decoder is similar to the decoder of GPT-Net. SegNet uses SGD optimizer, with a total of 3,500 iterations. Most categories in the dataset use the same probability to simulate structural and logical anomalies during the anomaly simulation phase. Each batch for training contains 4 normal images and 4 simulated abnormal images. The hyper-parameters of SegNet are as follows: the learning rate is 0.04; $\gamma$ in the focal loss is 4; $\alpha_4$, $\alpha_5$, and $\alpha_6$ in the loss function are 0.4, 0.6, and 0.3, respectively. As with the semantic segmentation network, SLSG directly gives the anomaly score for each pixel in the input image, and gives the image-level anomaly score by averaging 100 most abnormal pixel points.

### 4.3. Comparison with Other Methods

In this subsection, we compare SLSG with different methods. From the quantitative perspective, Tab. 1, 2, and 3 list the AUC scores of the different methods on the three benchmark datasets, respectively. For all three datasets in experiments, SLSG has the best ROC-AUC score at the image level among the models compared and achieves SOTA performance on the LOCO AD dataset, demonstrating the effectiveness of SLSG in anomaly detection. From the qualitative perspective, Fig. 5 shows the results of anomaly localization of the different methods. SLSG is able to determine whether each pixel in the image is abnormal or not with a higher confidence. The results of anomaly localization also have the smallest error with GTs, which is thanks to the pixel-level self-supervised task we designed.

Table 1. The performance for anomaly detection and localization of different methods on the BeanTech AD dataset with the format of (Image-level ROC-AUC%, Pixel-level ROC-AUC%).

| Category | SPADE [9] | PaDiM [10] | PatchCore [8] | P-SVDD [56] | Ours |
|---|---|---|---|---|---|
| 01 | (91.4,**97.3**) | (99.8,97.0) | (90.9,95.5) | (95.7,91.6) | (**100**,96.5) |
| 02 | (71.4,94.4) | (82.0,96.0) | (79.3,94.7) | (72.1,93.6) | (**85.4,96.3**) |
| 03 | (**99.9**,99.1) | (99.4,98.8) | (99.8,**99.3**) | (82.1,91.0) | (99.3,99.0) |
| Mean | (87.6,96.9) | (93.7,97.3) | (90.0,96.5) | (83.3,92.1) | (**94.9,97.3**) |

Meanwhile, in order to analyze the ability of SLSG to detect logical and structural anomalies respectively, we divide the samples in the test set of LOCO AD into three categories: normal, logical anomaly, and structural anomaly. As shown in Fig. 6, for the models that perform well on structural anomalies, such as PaDiM and PatchCore, due to the lack of the modeling of position relationships, they are less effective when used directly for the detection of logical anomalies. GCAD designs a two-branch structure that focuses on the detection of logical and structural anomalies respectively, but the detection performance for structural anomalies is limited. SLSG is designed efficiently from both encoding and decoding perspectives, and achieves a better balance in the detection of logical and structural anomalies.

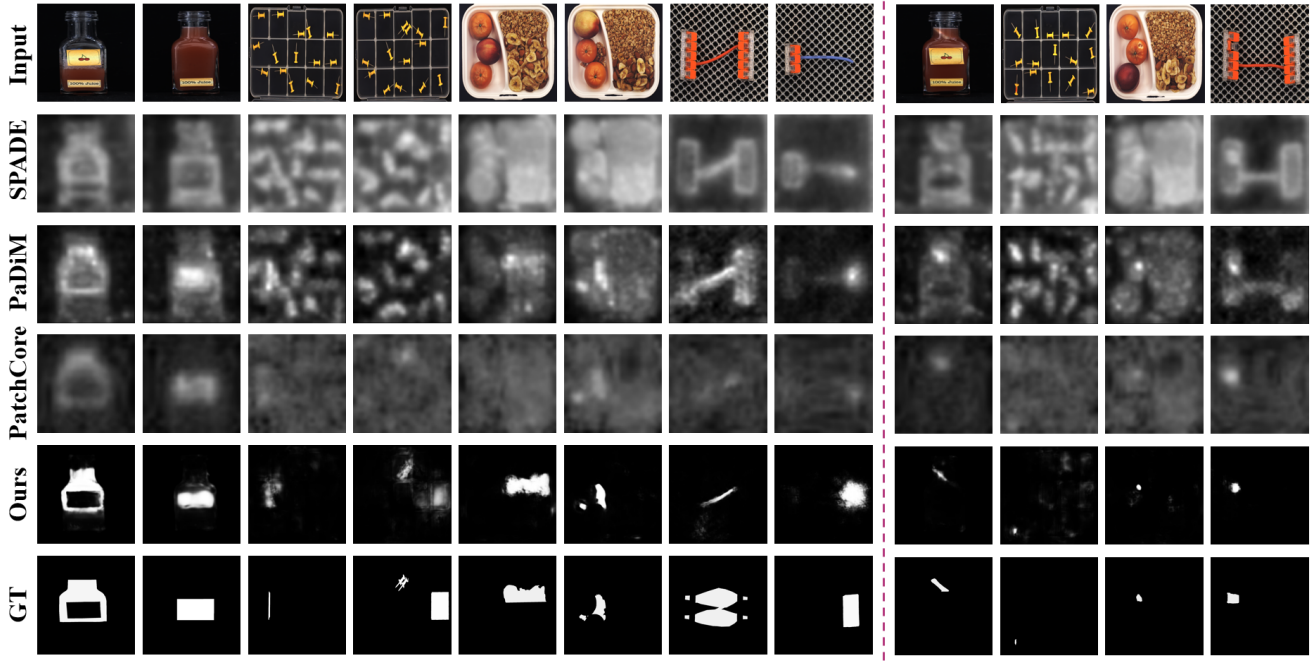

Figure 5. The result of anomaly localization of SPADE, PaDiM, PatchCore and our method on the MVTec LOCO AD dataset. Logical anomalies (left) and structural anomalies (right). SLSG gives anomaly discrimination with a higher confidence level.

Table 2. The performance for anomaly detection and localization of different methods on the MVTec AD dataset with the format of (Image-level ROC-AUC%, Pixel-level ROC-AUC%).

| | Category | SPADE [9] | PaDiM [10] | DRAEM [34] | CutPaste [32] | P-SVDD [56] | GCAD [17] | Ours |
|---|---|---|---|---|---|---|---|---|
| Texture | carpet | (-,97.5) | (-,98.9) | (97.0,95.5) | (92.9,92.6) | (93.1,98.3) | (-,-) | (**99.0,96.0**) |
| | grid | (-,93.7) | (-,94.9) | (99.9,**99.7**) | (94.6,96.2) | (99.9,97.5) | (-,-) | (**100**,98.5) |
| | leather | (-,97.6) | (-,99.1) | (100,98.6) | (90.9,97.4) | (100,99.5) | (-,-) | (**100,99.5**) |
| | tile | (-,87.4) | (-,91.2) | (99.6,99.2) | (97.8,91.4) | (93.4,90.5) | (-,-) | (**100,98.6**) |
| | wood | (-,85.5) | (-,93.6) | (99.1,96.4) | (96.5,90.8) | (98.6,95.5) | (-,-) | (**99.6,96.8**) |
| | average | (-,92.3) | (-,95.6) | (99.1,97.9) | (94.5,93.7) | (97.0,96.3) | (-,-) | (**99.7,97.9**) |
| Object | bottle | (-,98.4) | (-,98.1) | (99.2,99.1) | (98.6,98.1) | (98.3,97.6) | (-,-) | (**99.4,99.1**) |
| | cable | (-,97.2) | (-,95.8) | (91.8,94.7) | (90.3,96.8) | (80.6,90) | (-,-) | (**98.3**,97.4) |
| | capsule | (-,99.0) | (-,98.3) | (98.5,94.3) | (76.7,95.8) | (96.2,97.4) | (-,-) | (95.5,95.9) |
| | hazelnut | (-,99.1) | (-,97.7) | (100,**99.7**) | (92.0,97.5) | (97.3,97.3) | (-,-) | (99.5,97.8) |
| | metal nut | (-,98.1) | (-,96.7) | (98.7,**99.5**) | (94.0,98.0) | (99.3,93.1) | (-,-) | (**100**,98.9) |
| | pill | (-,96.5) | (-,94.7) | (98.9,97.6) | (86.1,95.1) | (92.4,95.7) | (-,-) | (**99.2,98.0**) |
| | screw | (-,**98.9**) | (-,97.4) | (93.9,97.6) | (81.3,95.7) | (86.3,96.7) | (-,-) | (89.1,97.3) |
| | toothbrush | (-,97.9) | (-,98.7) | (100,98.1) | (100,98.1) | (98.3,98.1) | (-,-) | (**100,99.4**) |
| | transistor | (-,94.1) | (-,97.2) | (93.1,90.9) | (91.5,97) | (95.5,93.0) | (-,-) | (**97.3**,92.5) |
| | zipper | (-,96.5) | (-,98.2) | (100,98.8) | (97.9,95.1) | (99.4,**99.3**) | (-,-) | (**100**,97.1) |
| | average | (-,97.57) | (-,97.3) | (97.4,97.0) | (90.8,96.7) | (94.3,95.8) | (-,-) | (**97.8,97.3**) |
| | Mean | (85.5,96.0) | (95.3,96.7) | (98.0,97.3) | (95.2,96.0) | (92.1,95.7) | (93.1,-) | (**98.5,97.5**) |

Table 3. The performance for anomaly detection and localization of different methods on the MVTec LOCO AD dataset with the format of (Image-level ROC-AUC%, Pixel-level sPRO-AUC%).

| Category | f-AnoGAN [57] | AE [17] | S-T [20] | SPADE [9] | PaDiM [10] | PatchCore [8] | GCAD [17] | Ours |
|---|---|---|---|---|---|---|---|---|
| Breakfast box | (60.1,22.3) | (52.8,18.9) | (68.6,49.6) | (78.2,37.2) | (65.7,47.7) | (81.3,46.0) | (83.9,50.2) | (**88.9,65.9**) |
| Juice bottle | (80.1,56.9) | (65.2,60.5) | (91.0,81.1) | (88.3,80.4) | (88.9,82.4) | (95.6,71.0) | (**99.4,91.0**) | (99.1,82.0) |
| Pushpins | (66.9,33.6) | (64.1,32.7) | (74.9,52.3) | (59.3,23.4) | (61.2,34.8) | (72.3,44.7) | (86.2,73.9) | (**95.5,74.4**) |
| Screw bag | (47.9,34.8) | (44.1,28.9) | (71.2,60.2) | (53.2,33.1) | (60.9,47.2) | (64.9,52.2) | (63.2,**55.8**) | (**79.4**,47.2) |
| Splicing connector | (66.3,19.5) | (60.4,47.9) | (81.1,69.8) | (65.4,51.6) | (67.8,47.6) | (82.4,58.6) | (83.9,**79.8**) | (**88.5**,66.9) |
| Mean | (64.2,33.4) | (57.3,37.8) | (77.3,62.6) | (68.8,45.1) | (68.9,52.0) | (79.3,54.5) | (83.3,**70.1**) | (**90.3**,67.3) |

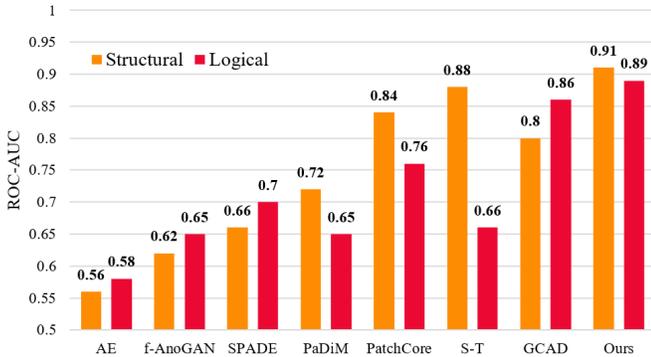

Figure 6. Comparison of our model with different models for the detection of structural and logical anomalies.

## 4.4 Effect of Pre-Training Strategy

The GPT-Net improves the feature embedding ability of the encoder and enhances the sensitivity of the encoder to position information. In the pre-training stage, GPT-Net uses grid mask and large mask to complete the mask operation on the image.

Table 4. Image-level ROC-AUC scores of SLSG on the LOCO AD dataset when using different pre-training strategies.

|  | 8-60% | 8-90% | 4-80% | 16-80% | No | ImageNet | SLSG |
|---|---|---|---|---|---|---|---|
| Structural | 0.894 | 0.883 | 0.849 | 0.875 | 0.911 | 0.846 | **0.914** |
| Logical | 0.884 | 0.887 | 0.866 | 0.854 | 0.856 | 0.755 | **0.896** |
| Mean | 0.886 | 0.881 | 0.851 | 0.858 | 0.877 | 0.792 | **0.903** |

For the grid mask, GPT-Net uses patches of size 8×8 to randomly mask 80% of the area (8-80%). Tab. 4 shows the influences of different mask sizes and mask ratios on anomaly detection. Overall, the size of the mask has a greater effect on anomaly detection, while the ratio of the mask within a certain range has less effect. For the large mask, which further enhances the ability of the encoder to model remote relationships, when it is not used (column "No" in Tab. 4), the detection performance of SLSG for logical anomalies drops significantly. Also, Tab. 4 compares the impact of the generation-based pre-training strategy and the ImageNet-based pre-training strategy (column "ImageNet" in Tab. 4) on SLSG, as described in 3.1, the generation-based pre-training strategy is better at feature embedding.

## 4.5 Effect of Anomaly Simulation Strategy

To better classify normal patterns, SLSG builds a self-supervised task by introducing simulated abnormal samples. For the anomaly simulation strategy, SLSG mainly uses three strategies, structure anomaly simulation, logical anomaly simulation, and foreground enhancement strategy. The ablation experiments in Tab. 5 demonstrate the effectiveness of these three strategies. When logical anomaly simulation is not used, the ROC-AUC score of SLSG decreases by 3.8% for logical anomaly detection and by 0.8% for structural anomaly detection. This indicates that these two anomaly simulation strategies not only have a specific boost to the corresponding task, but also to additional task by working in conjunction with each other. In addition, the shape of the simulated anomalies created by SLSG with the help of Perlin noise is much closer to the shape of the real anomalies. To measure the effect of different shapes of noise on anomaly detection, we generate simulated abnormal images using rectangular noise (Rect. Noise) instead of Perlin noise, but the ROC-AUC score of the model for anomaly detection decreased by only 2.5%. This demonstrates that even if the shape of the simulated anomalies during training is only rectangle, the model can easily generalize to the detection of real anomalies with irregular shapes during the inference stage. Further, it demonstrates the effectiveness of the method that using self-supervised tasks to assist semi-supervised anomaly detection.

Table 5. Image-level ROC-AUC scores of SLSG on the LOCO AD dataset when using different anomaly simulation strategies.

|  | w/o Structural | w/o Logical | w/o Foreground | Rect. Noise | SLSG |
|---|---|---|---|---|---|
| Structural | 0.850 | 0.906 | 0.908 | 0.896 | **0.914** |
| Logical | 0.876 | 0.858 | 0.886 | 0.881 | **0.896** |
| Mean | 0.861 | 0.880 | 0.892 | 0.878 | **0.903** |

In Section 1, we mentioned that one-class classifier establishes a closed decision boundary by modeling normal samples, and treats samples outside the decision boundary as non-normal. Therefore, it is more suitable for the task of semi-supervised anomaly detection. Theoretically, combined with the context that the products produced on the same product line are highly consistent, we elaborate on the learning principle of the SLSG to implement the one-class classifier. Experimentally, to further verify whether the decoder of SegNet achieves the effect of one-class classification, we visualize the distribution of the feature maps in two-dimensional space in Fig. 7 using t-SNE [58]. The feature map visualized comes from the output of the first block of the decoder, it contains some information for identifying anomalies after the preliminary decoding.

For each subfigure in Fig. 7, we specifically visualize the simulated abnormal samples generated during training, all normal and real abnormal samples in the test set. We also mark the samples misclassified by SLSG during the inference stage and the approximate decision boundaries. From the results of the visualization, there are three main conclusions: (1) The distribution of some simulated abnormal samples coincides with the distribution of real abnormal samples. This shows that simulated abnormal samples can act as real abnormal samples, demonstrating the effectiveness of the anomaly simulation strategy; (2) The distribution of normal samples is centralized and the distribution of abnormal samples is more scattered. This demonstrates that the features of the normal samples are easy to generalize, while the features of the abnormal samples are diverse and difficult to unify; (3) Combined with the training process of the model (Fig. 7-h), it can be argued that by comparing highly consistent normal samples and variable simulated abnormal samples, SLSG constructs a one-class classification decision boundary. Using this decision boundary, SLSG can better distinguish between normal samples and unseen non-normal samples.

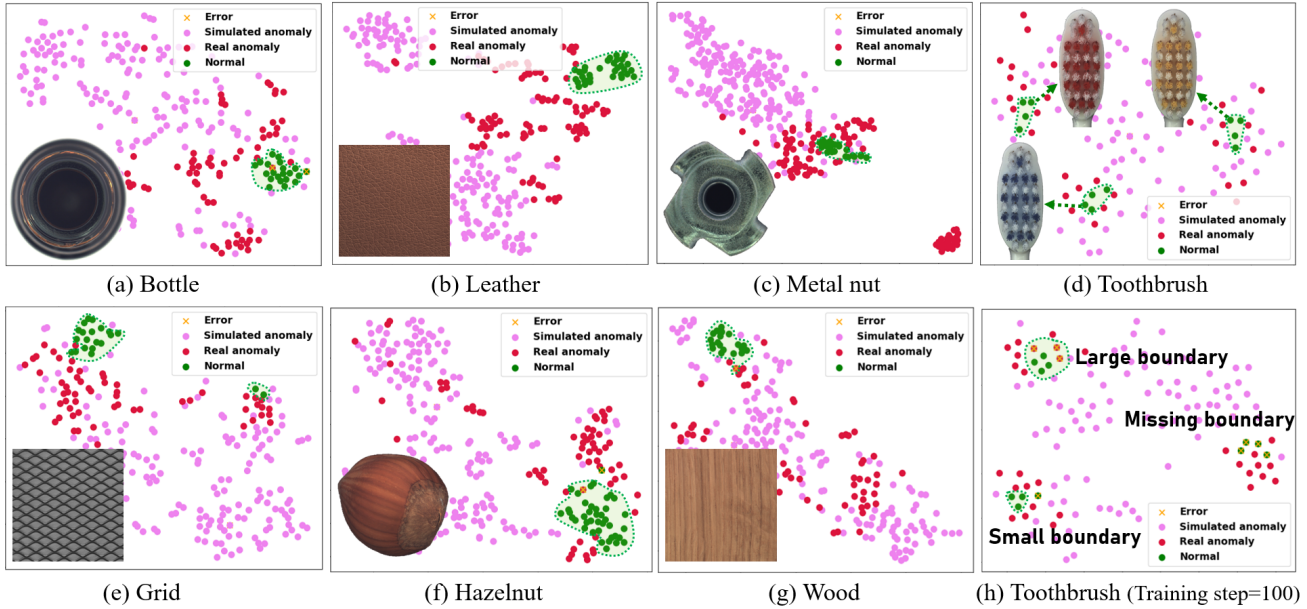

Figure 7. The distribution of simulated abnormal samples, real abnormal samples, and normal samples in two-dimensional space. The approximate decision boundaries drawn show that SLSG implements the one-class classification by learning.

### 4.6 Effect of SG Block

For SLSG, an important module is the SG block, which models both extensive and key logical relationships effectively. In Tab. 6, we use the LOCO AD dataset to conduct ablation experiments on important structures and parameters of the SG block. We use U-Net [38] as the baseline. Experiment 2 demonstrates that adding additional skip connections to the decoder from the bottleneck structure facilitates the detection of logical anomalies, while it does not facilitate the detection of structural anomalies. Experiments 3 and 4 respectively report the performance of using only self-attention network and only GCN network, both of which can improve the detection performance of the model to logical anomalies when used alone.

At the same time, while constructing the graph structure, the sparse similarity matrix $W_s$ used by SLSG eliminates the position similarity (PS), which can make the model pay more attention to the logical relationship among cross-neighborhood nodes. When the position similarity is not eliminated (Experiment 5), the performance of model to detect structural anomalies is almost unchanged, but the performance to detect logical anomalies decreases. In addition, for each node in the graph, SLSG selects 9 nodes that are most similar to it as its neighboring nodes. The experiments $SLSG_{k=7,9,11}$ in Tab. 6 demonstrate that different numbers of neighboring nodes have a limited influence on the model.

To further validate the modeling effectiveness of the SG block, we visualize the graph structure created by it. In Fig. 8, the patch marked with green dot is the central node and the patches marked with red dots are its neighboring nodes. It can be seen that the neighboring nodes selected by SG block have a high visual and semantic similarity with the central node, and the selection of neighboring nodes is not limited to the first-order neighborhoods of central node. This further proves the effectiveness of the SG block.

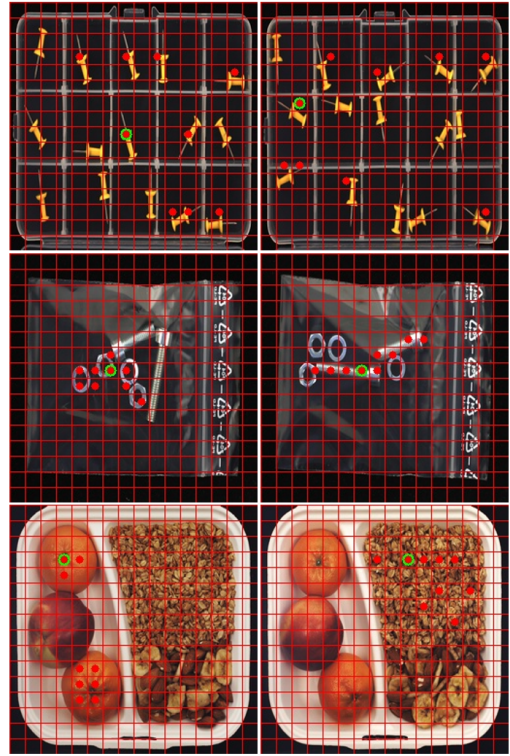

Figure 8. Graph structure constructed by GCN.

### 4.7 Effect of Different Loss Functions

SegNet uses $\ell 1$ loss and focal loss to calculate the distance between predicted and true values. Also, SegNet introduces auxiliary loss at the bottleneck structure to directly optimize the parameters of the SG block. Tab. 7 reports the average ROC-AUC scores of different loss functions for different anomaly types. In the training stage, SLSG accomplishes the semantic segmentation task of binary classification. However, since the areas of anomalies in simulated abnormal images are generally small, the problem of data imbalance at the pixel level makes it hard for the model to converge only with the constraint of $\ell1$ loss. Therefore, SLSG introduces focal loss to ease the training difficulty. On the other hand, focal loss is more tolerant of small errors between predicted and true values, while $\ell1$ loss is more sensitive to small errors. So, it is difficult for SLSG to give high-confidence anomaly judgements with focal loss alone (as shown in Fig. 9-b), while introducing $\ell1$ loss on the basis of focal loss can alleviate the blurring problem of segmented images. Also, by introducing the auxiliary loss, SLSG achieves a 2% performance improvement in the detection of logical anomalies.

Table 6. The image-level ROC-AUC scores of SLSG when using different modules and parameters.

| Experiment | Bottleneck Skip | Self-Attention | GCN | Eliminate PS | TopK=7 | TopK=9 | TopK=11 | Structural | Logical | Mean |
|---|---|---|---|---|---|---|---|---|---|---|
| 1 | | | | | | | | 0.899 | 0.846 | 0.867 |
| 2 | √ | | | | | | | 0.875 | 0.865 | 0.868 |
| 3 | √ | √ | | | | | | 0.883 | 0.876 | 0.875 |
| 4 | √ | | √ | √ | | √ | | 0.876 | 0.885 | 0.877 |
| 5 | √ | √ | √ | | | √ | | 0.912 | 0.882 | 0.889 |
| $SLSG_{k=7}$ | √ | √ | √ | √ | √ | | | **0.915** | 0.885 | 0.894 |
| $SLSG_{k=9}$ | √ | √ | √ | √ | | √ | | 0.914 | **0.896** | **0.903** |
| $SLSG_{k=11}$ | √ | √ | √ | √ | | | √ | 0.906 | **0.896** | 0.898 |

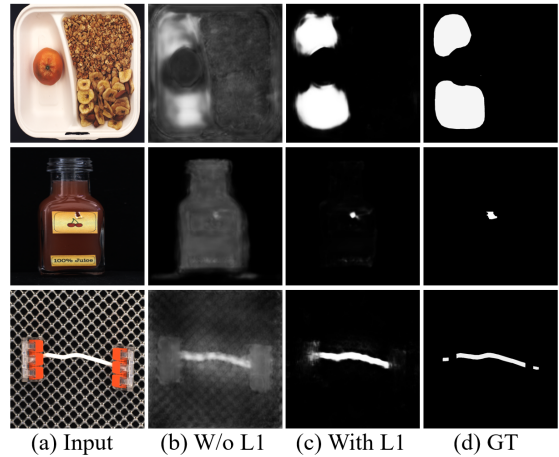

(a) Input  (b) W/o L1  (c) With L1  (d) GT

Figure 9. Effect of different loss functions on anomaly localization.

Table 7. Evaluation of different loss functions. The image-level ROC-AUC scores on the LOCO AD dataset are reported for different loss functions.

| L1 loss | Focal loss | Aux loss | Structural | Logical | Mean |
|---|---|---|---|---|---|
| | √ | √ | 0.885 | 0.868 | 0.872 |
| √ | | √ | 0.873 | 0.875 | 0.868 |
| √ | √ | | **0.916** | 0.876 | 0.891 |
| √ | √ | √ | 0.914 | **0.896** | **0.903** |

In the pre-training stage of SLSG, the inputs to GPT-Net lose a lot of detailed information after the large percentage of mask operation. In this case, requiring the model to predict this detailed information is demanding. As shown in Fig. 10-d, in the case of using only $\ell1$ loss, GPT-Net will excessively pursue the pixel-by-pixel consistency of the reconstructed image with the ground truth. This results in the clear reconstructions of unmasked regions and the blurred reconstructions of masked regions. Compared with $\ell1$ loss, the constraint of adversarial loss is relatively weak, and it only requires the distribution of the reconstructed image and the ground truth to be consistent at the global scale [59]. Therefore, introducing the adversarial loss in the pre-training stage can alleviate the blurring problem of the reconstructed images, generate clearer and more realistic textures, and enhances the feature representation ability of the encoder.

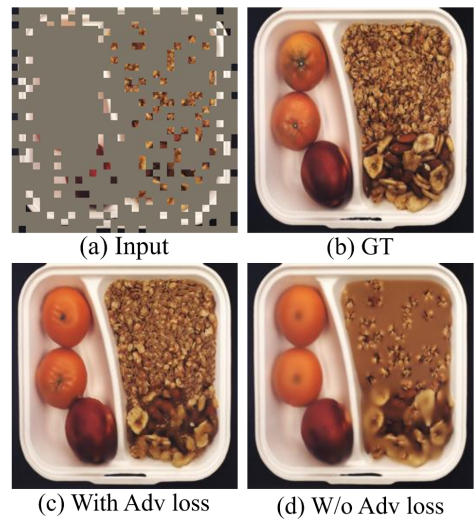

(a) Input  (b) GT

(c) With Adv loss  (d) W/o Adv loss

Figure 10. Effect of different loss functions on image reconstruction during the pre-training stage.

### 4.8 Computational Efficiency

SLSG is primarily focused on the anomaly detection of product surfaces in industrial scenarios. In order to meet the efficiency requirements of industrial production, real-time performance is also a key factor in addition to detection accuracy. We compare the inference speed of SLSG with feature representation-based models, which generally have high anomaly detection accuracy, on an NVIDIA GTX1080Ti. To simulate online detection, we set the batch size in the inference stage to 1. The average time to detect an image for SPADE-WR50-Top1 [9] and PaDiM-R18-Rd100 [10] is 480 ms and 471 ms respectively, while SLSG only takes 54.2 ms. For offline detection, i.e. when the batch size is set to 16, the average inference time of SLSG for an image is only 18.5 ms. In addition, the total number of parameters in SLSG is only 20.46 MB, the memory usage in the inference stage is 286.95 MB, and the theoretical amount of floating-point arithmetic is 23 GFlops. Thus, SLSG has lower requirements for hardware devices, which is conducive to its low-cost deployment in industrial scenarios.

### 4.9 Limitations

When using the LOCO AD dataset for anomaly detection, SLSG achieves SOTA performance on ROC-AUC scores at the image level, but sub-optimal performance on sPRO-AUC scores at the pixel level. Taking juice bottle as an example, as shown in Tab. 3, the anomaly detection performance of SLSG at the image level is much better than PaDiM (0.991 vs 0.889), but the anomaly localization performance at the pixel level is slightly worse (0.820 vs 0.824). Further, we visualize the variation of FPR and sPRO versus binary threshold for these two models on juice bottles in Fig. 11.

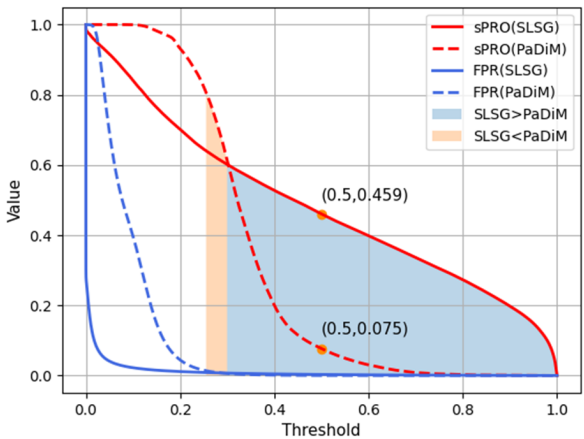

Figure 11. FPR and sPRO values of PaDiM and SLSG at different thresholds.

In Fig. 11, as the threshold increases, the FPR of SLSG decreases rapidly and sPOR decreases steadily in a linear trend. However, the FPR of PaDiM decreases slowly and sPRO has a high sensitivity within a certain threshold interval. For the optimal threshold 0.5 of the binary classification, the FPR of both models is close to 0, but the sPRO of SLSG is much larger than PaDiM (0.459 vs 0.075). Also, all metrics of SLSG outperform PaDiM in the 70% threshold interval (the blue-filled area), while the interval in which all metrics of PaDiM outperform SLSG is only 5% (the orange-filled area). Combined with Fig. 5 and 11, it is demonstrated that although the sPRO-AUC score of SLSG is worse than some models, its end-to-end anomaly segmentation approach is able to complete anomaly discrimination with a high confidence level. Therefore, the FPR and sPRO of SLSG are less affected by the binary threshold, which ensures the robustness of SLSG in practical applications.

In addition, SLSG has limitations in modeling the multi-object logical relationships. For example, when three oranges appear in the breakfast box, the model will have the judgment of false negatives. To address this shortcoming, we can try to model the pattern of normal samples more comprehensively using better model frameworks and constraints.

## 5. Conclusion

In this paper, we propose a novel method called SLSG to address semi-supervised image anomaly detection. SLSG uses a generative pre-trained network to learn the feature embedding of normal images. Also, it uses simulated abnormal images to provide pseudo-prior knowledge of anomalies, thereby trains an end-to-end one-class classifier with a compact decision hypersphere to complete anomaly detection. Meanwhile, with the help of reasoning mask regions and constructing a general graph structure, SLSG can capture the long-range dependencies among elements in the image. Extensive experiments have demonstrated the comprehensive and superior anomaly detection performance of SLSG. In addition, thanks to the end-to-end anomaly discrimination and fully convolutional structure, the computational cost of SLSG is also smaller.